\pdfoutput=1

\documentclass[11pt]{article}

\usepackage{emnlp2021}

\usepackage{times}
\usepackage{latexsym}

\usepackage[T1]{fontenc}

\usepackage[utf8]{inputenc}

\usepackage{microtype}

\usepackage{booktabs}       
\usepackage{amsfonts}       
\usepackage{nicefrac}       
\usepackage{caption}
\usepackage{subcaption}
\usepackage{graphicx}
\usepackage{amsmath, amsmath}
\usepackage{multirow}
\usepackage{lipsum}
\usepackage{breqn}
\usepackage{empheq}
\usepackage{multirow}
\usepackage{bbm}
\def\ie{{\em i.e.,~}}

\usepackage{xcolor}
\newcommand{\dejiao}[1]{{\color{black}{#1}}}
\usepackage{todonotes}
\usepackage{hyperref}
\definecolor{darktaupe}{rgb}{0.28, 0.24, 0.2}

\usepackage{adjustbox}
\usepackage{array}
\usepackage{color, colortbl}

\newcolumntype{R}[2]{%
    >{\adjustbox{angle=#1,lap=\width-(#2)}\bgroup}%
    l%
    <{\egroup}%
}

\title{Pairwise Supervised Contrastive Learning of Sentence Representations}

\author{Dejiao Zhang \quad Shang-Wen Li\thanks{$^*$Equal contribution. Correspondence to Dejiao Zhang <dejiaoz@amazon.com>. This work is accepted to EMNLP 2021 and our code is released at \url{https://github.com/amazon-research/sentence-representations}} \quad Wei Xiao$^*$ \quad  Henghui Zhu$^*$ \\ \quad   \quad
\textbf{Ramesh Nallapati} \quad 
\textbf{Andrew O. Arnold}\quad \textbf{Bing Xiang} \\
  AWS AI
}
  
\date{}

\begin{document}
\maketitle

\begin{abstract}
Many recent successes in sentence representation learning have been achieved by simply fine-tuning on the Natural Language Inference (NLI) datasets with triplet loss or siamese loss. Nevertheless, they share a common weakness: sentences in a contradiction pair are not necessarily from different semantic categories. Therefore, optimizing the semantic entailment and contradiction reasoning objective alone is inadequate to capture the high-level semantic structure. The drawback is compounded by the fact that the vanilla siamese or triplet losses only learn from individual sentence pairs or triplets, which often suffer from bad local optima. In this paper, we propose PairSupCon, an instance discrimination based approach aiming to bridge semantic entailment and contradiction understanding with high-level categorical concept encoding. We evaluate PairSupCon on various downstream tasks that involve understanding sentence semantics at different granularities. We outperform the previous state-of-the-art method with $10\%$--$13\%$ averaged improvement on eight clustering tasks, and $5\%$--$6\%$ averaged improvement on seven semantic textual similarity (STS) tasks.   
\end{abstract}

\section{Introduction}
\label{sec:introduction}
Learning high-quality sentence embeddings is a fundamental task in Natural Language Processing.  The goal is to map semantically similar sentences close together and dissimilar sentences farther apart in the representation space. Many recent successes have been achieved by training on the NLI datasets \citep{bowman2015large, williams2017broad,wang2018glue}, where the task is often to classify each sentence pair into one of three categories: entailment, contradiction, or neutral.
\begin{figure}[htbp]
    \centering
    \begin{subfigure}{0.46\textwidth}
        \centering
         \includegraphics[scale=0.33]{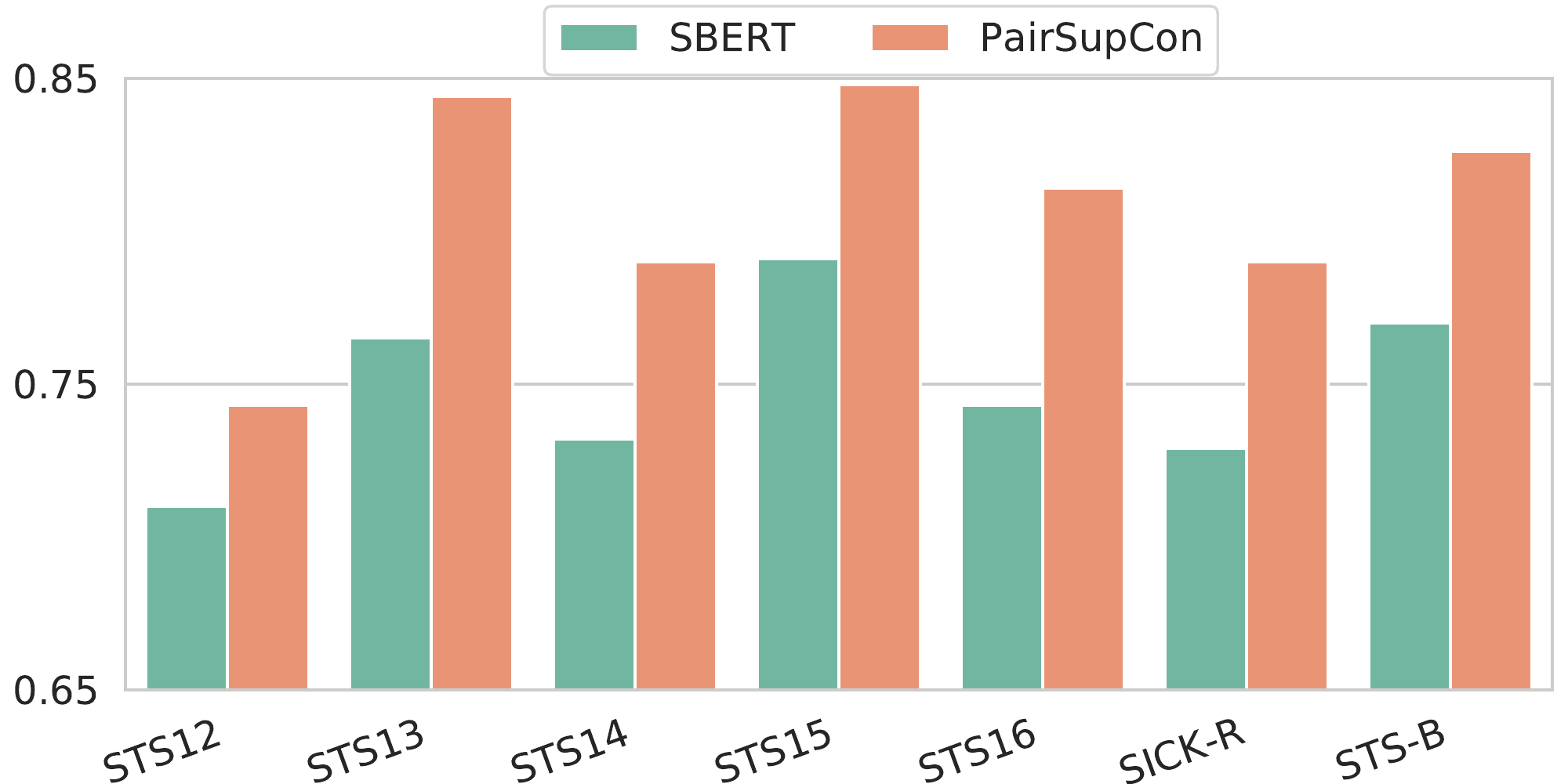}
        \caption{Spearman Correlation Evaluation on STS.}
        \label{fig:stshist}
    \end{subfigure}
    \vspace{0.2cm}
    
     \begin{subfigure}{0.46\textwidth}
        \centering
        \includegraphics[scale=0.31]{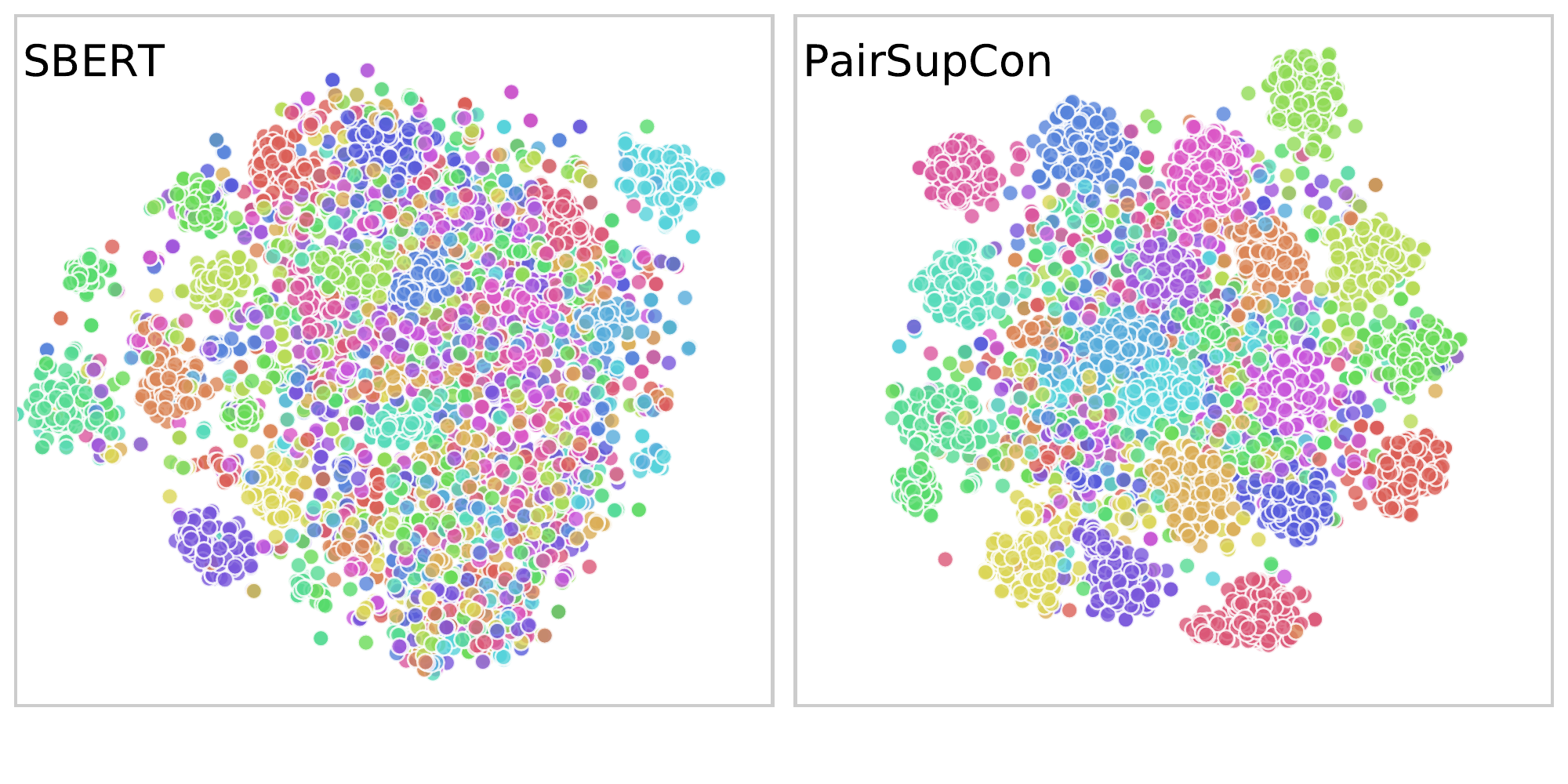}
        \caption{TSNE visualization of the StackOverflow embeddings.}
        \label{fig:tsne_stackoverflow}
    \end{subfigure}
    \caption{Simultaneoulsy encoding semantic categorical structure and pairwise entailment and contradiction understanding into embeddings. }
    \label{fig:motivation_validation}
\end{figure}

Despite promising results, prior work \citep{conneau2017supervised, cer2018universal, reimers-2019-sentence-bert} share a common weakness: the sentences forming a contradiction pair may not necessarily belong to different semantic categories. Consequently, optimizing the model for semantic entailment and contradiction understanding alone is inadequate to encode the high-level categorical concepts into the representations. Moreover, the vanilla siamese (triplet) loss only learns from the individual sentence pairs (triplets), which often requires substantial training examples to achieve competitive performance \citep{oh2016deep,thakur2020augmented}. 
As shown in Section \ref{subsec:clustering}, the siamese loss can sometimes drive a model to bad local optima where the performance of high-level semantic concept encoding is degraded when compared with its baseline counterpart.

In this paper, we take inspiration from self-supervised contrastive learning \citep{bachman2019learning,he2020momentum, chen2020simple} and propose jointly optimizing the pairwise semantic reasoning objective with an \textit{instance discrimination loss}. We name our approach Pairwise Supervised Contrastive Learning (PairSupCon). \dejiao{As noticed by the recent work \citep{wu2018unsupervised, zhang-etal-2021-supporting}, instance discrimination learning can implicitly group similar instances together in the representation space without any explicit learning force directs to do so.}  PairSupCon leverages this implicit grouping effect to bring together representations from the same semantic category while, simultaneously enhancing the semantic entailment and contradiction reasoning capability of the model. 

Although the prior work mainly focuses on pairwise semantic similarity related evaluations, we argue in this paper that the capability of encoding the high-level categorical semantic concept into the representations is an equally important aspect for evaluations. As shown in Section \ref{sec:experiments}, the previous state-of-the-art model that performs best on the semantic textual similarity (STS) tasks results in degenerated embeddings of the categorical semantic structure. On the other hand, better capturing the high-level semantic concepts can in turn promote better performance of the low-level semantic entailment and contradiction reasoning. This assumption is consistent with how human categorize objects in a top-down reasoning manner. 
We further validate our assumption in Section \ref{sec:experiments}, where PairSupCon achieves an averaged improvement of $10\%-13\%$ over the prior work when evaluated on eight short text clustering tasks, and yields $5\%-6\%$ averaged improvement on seven STS tasks.

\section{Related Work}
\paragraph{Sentence Representation Learning with NLI}{
The suitability of leveraging NLI to promote better sentence representation learning is first observed by InferSent \citep{conneau2017supervised}, where a siamese BiLSTM network is optimized in a supervised manner with the semantic entailment and contraction classification objective. Universal Sentence Encoder \citep{cer2018universal} later augments an unsupervised learning objective with the supervised learning on NLI, and shows better transfer performance on various downstream tasks. 

More recently, SBERT \citep{reimers2019sentence} finetunes a siamese BERT \citep{devlin2018bert} model on NLI and sets new state-of-the-art results. However, SBERT as well as the above work adopt the vanilla siamese or triplet loss, which often suffer from slow convergence and bad local optima \citep{oh2016deep,thakur2020augmented}. 
}

\paragraph{Self-Supervised Instance Discrimination}{
Another relevant line of work is self-supervised contrastive learning, which essentially solves an instance discrimination task that targets at discriminating each positive pair from all negative pairs within each batch of data \citep{oord2018representation,bachman2019learning,he2020momentum, chen2020simple}. 
Owing to their notable successes, self-supervised instance discrimination has become a prominent pre-training strategy for providing effective representations for a wide range of downstream tasks.

While recent successes are primarily driven by the computer vision domain, there is an increasing interest in leveraging variant instance discrimination tasks to support Pretraining Language Models (PLMs) \citep{meng2021coco,giorgi2020declutr,wu2020clear,rethmeier2021primer}. 
Our proposal can be seen as complementary to this stream of work, considering that the instance discrimination learning based PLMs provide a good foundation for PairSupCon to further enhance the sentence representation quality by further learning on NLI. 
As demonstrated in Section \ref{sec:experiments}, by training a pre-trained BERT-base model for less than an hour, PairSupCon attains substantial improvement on various downstream tasks that involve sentence semantics understanding at different granularities.
}

\begin{figure*}
    \centering
    \begin{subfigure}{0.49\textwidth}
        \centering
         \includegraphics[scale=0.40]{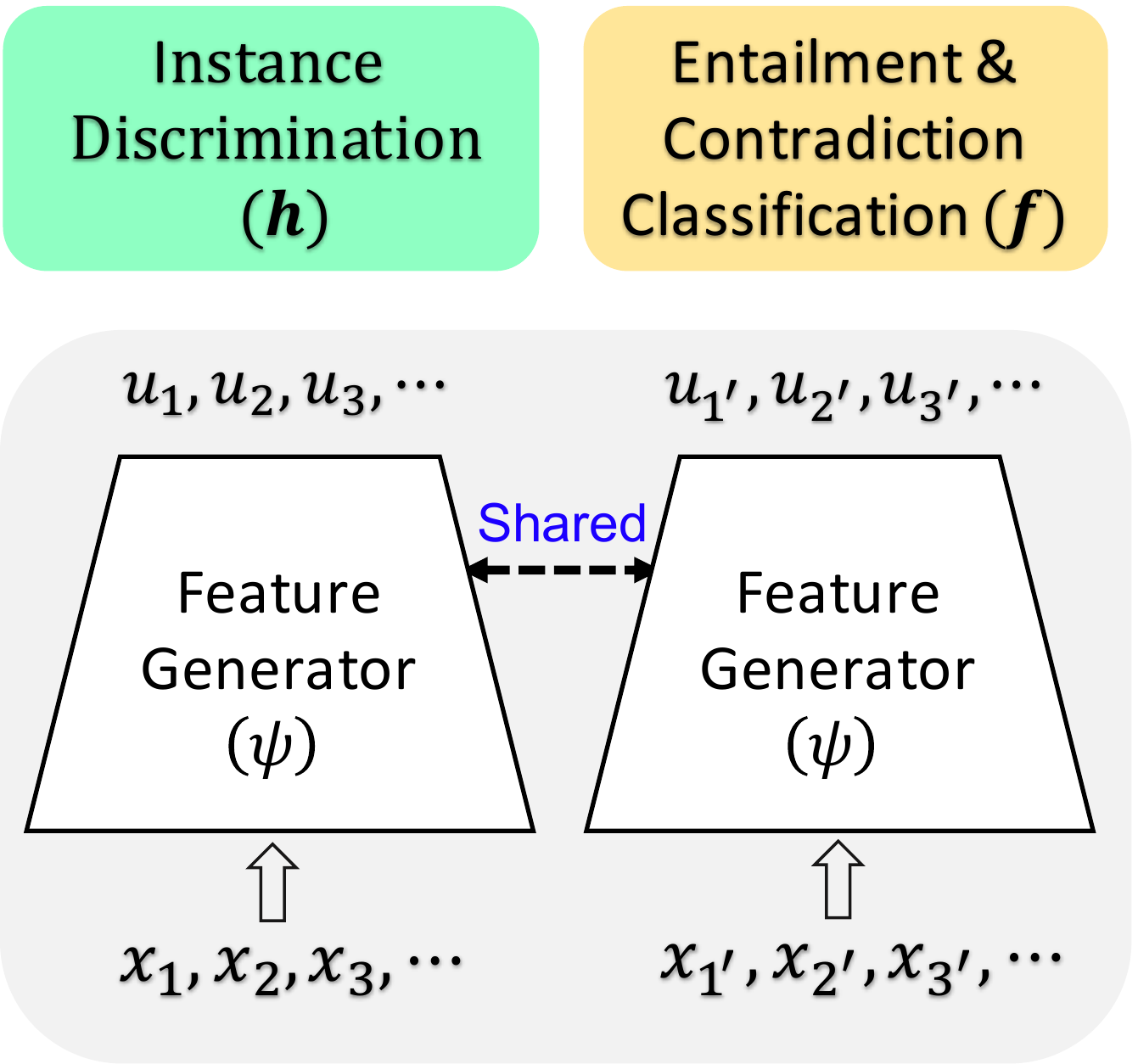}
    \caption{Joint optimization framework of PairSupCon.}
    \end{subfigure}
    \begin{subfigure}{0.49\textwidth}
        \centering
         \includegraphics[scale=0.39]{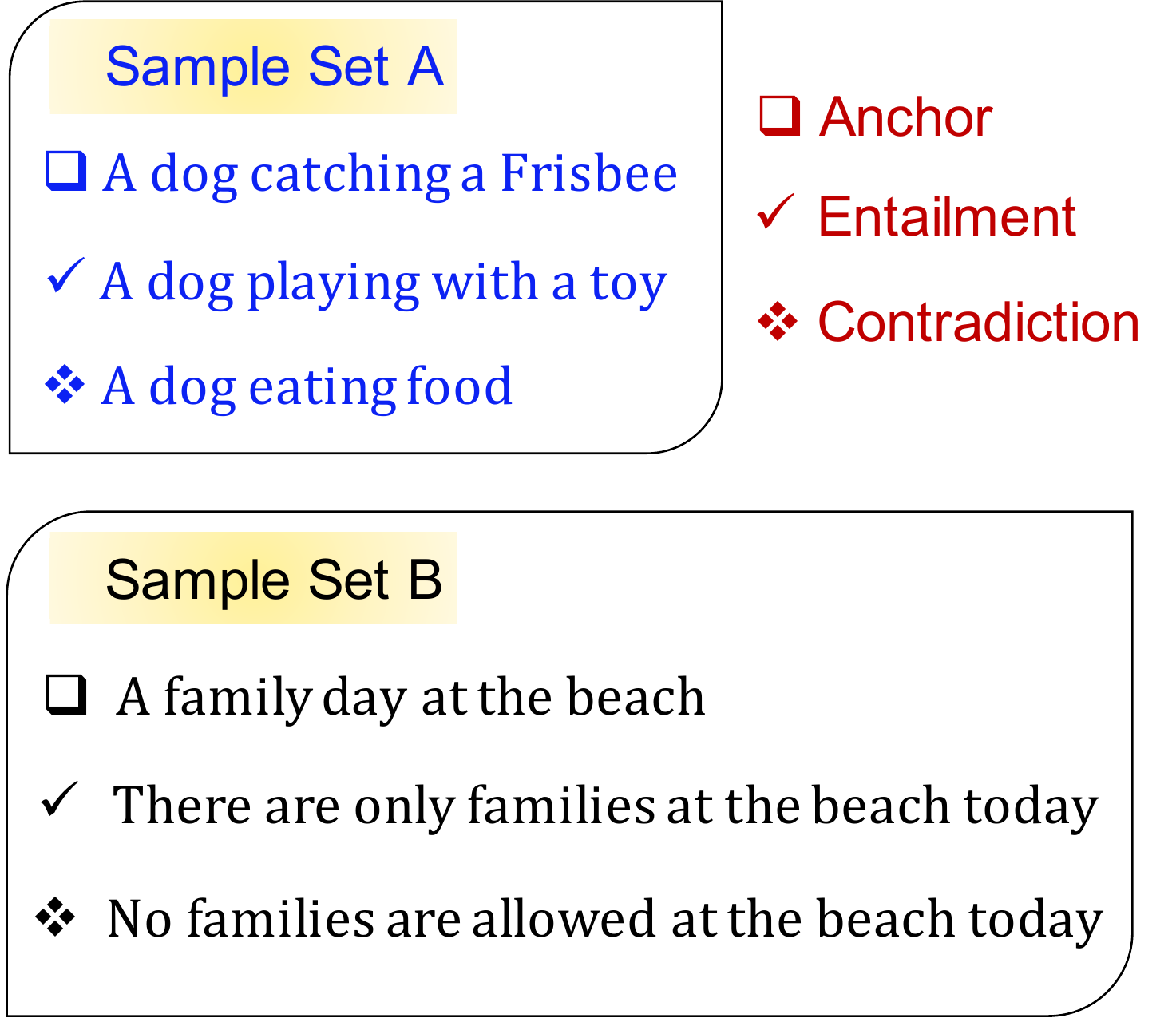}
    \caption{Entailment and contradiction samples.}
    \end{subfigure}
    \caption{Illustration of PairSupCon. \textbf{(a):} Training framework of PairSupCon. \textbf{(b)}: Despite the entailment sample is more similar to the anchor, both the contradiction and entailment samples are likely from the same semantic category as the anchor.}
    \label{fig:pairsupcon_model}
\end{figure*}

\paragraph{Deep Metric Learning}{
Inspired by the pioneering work of \citep{hadsell2006dimensionality,weinberger2009distance}, many recent works have shown significant benefit in learning deep representations using either siamese loss or triplet loss. However, both losses learn from individual pairs or triplets, which often require substantial training data to achieve competitive performance. Two different streams of work have been proposed to tackle this issue, with the shared focus on nontrivial pairs or triplets optimization. \citet{wang2014learning,schroff2015facenet,wu2017sampling,harwood2017smart} propose hard negative or hard positive mining that often requires expensive sampling. \citet{oh2016deep} extends the vanilla triplet loss by contrasting each positive example against multiple negatives. 

Our work leverages the strength of both lines, with the key difference being the above work requires categorical level supervision for selecting hard negatives. To be more specific, negative samples that have different categorical labels from the anchor but are currently mapped close to the anchor in the representation space, are likely to be more useful and hence being sampled. However, there are no categorical labels available in NLI. We thereby contrast each positive pair against multiple negatives collected using an unsupervised importance sampling strategy, for which the hypothesis is that hard negatives are more likely to locate close to the anchor. The effectiveness of this assumption is investigated in Section \ref{subsec:ablation}. 
}

\section{Model}
Following SBERT \citep{reimers-2019-sentence-bert}, we adopt the SNLI \citep{bowman2015large} and MNLI \citep{williams2017broad} as our training data, and refer the combined data as NLI for convenience. The NLI data consists of labeled sentence pairs and each can be presented in the form: (\textit{premise, hypothesis, label}). The premise sentences are selected from existing text sources and each premise sentence is paired with variant hypothesis sentences composed by human annotators. Each label indicates the hypothesis type and categorizes semantic relation of the associated premise and hypothesis sentence pair into one of three categories: \textit{entailment, contradiction}, and \textit{neural}, correspondingly.  

Prior work solely optimizes either siamese loss or triplet loss on NLI. We instead aiming to leverage the implicit grouping effect of instance discrimination learning to better capture the high-level categorical semantic structure of data  while, simultaneously promoting better convergence of the low-level semantic textual entailment and contradiction reasoning objective.

\subsection{Instance Discrimination}
We leverage the \textit{positive (entailment)} pairs of NLI to optimize an instance discrimination objective which tries to separate each positive pair apart from all other sentences. Let $\mathcal{D} = \left\{(x_j, x_{j'}), y_j\right\}_{j=1}^{M}$ denote a randomly sampled minibatch, with $y_i= \pm 1$ indicating an entailment or contradiction pair.  
Then for a premise sentence $x_i$ within a \textit{positive} pair $(x_i, x_{i'})$, we aim to separate its hypothesis sentence $x_i'$ from all other $2M$-2 sentences within the same batch $\mathcal{D}$. To be more specific, let $\mathcal{I} = \left\{i, i'\right\}_{i=1}^{M}$ denote the corresponded indices of the sentence pairs in $\mathcal{D}$, we then minimize the following for $x_i$,
\begin{align}
    \ell_{\text{ID}}^i = -\log \frac{\exp(s(z_i, z_{i'})/\tau)}{\sum_{j \in \mathcal{I}\setminus i} \exp(s(z_{i}, z_j)/\tau)}\;. 
    \label{eq:pairsupcon_pos}
\end{align}
In the above equation, $z_j = h(\psi(x_j))$ denotes the output of the instance discrimination head in Figure \ref{fig:pairsupcon_model}, $\tau$ denotes
the temperature parameter, and $s(\cdot)$ is chosen 
as the cosine similarity, \ie $\text{s}(\cdot) = z_i^Tz_{i'} / \|z_i\| \|z_{i'}\|$.  Notice that Equation (\ref{eq:pairsupcon_pos}) can be interpreted as a ($2M$--1)-way softmax based classification loss of classifying $z_i$ as $z_i'$. 

Similarly, for the hypothesis sentence $(x_{i'})$ we also try to discriminate its premise $(x_{i})$ from all the other sentences in $\mathcal{D}$. We denote the corresponding loss as $\ell_{\text{ID}}^{i'}$ that is defined by exchanging the roles of instances $x_{i'}$ and $x_{i}$ in Equation \eqref{eq:pairsupcon_pos}, respectively.    In summary, the final instance discrimination loss is averaged over all positive pairs in $\mathcal{D}$,  
\begin{align}
    \mathcal{L}_{\text{ID}} = \frac{1}{P_M}\sum_{i=1}^{M} \mathbbm{1}_{y_i=1} \cdot \left(\ell_{\text{ID}}^{i} + \ell_{\text{ID}}^{i'}\right)\;.
    \label{eq:pairsupcon_batchloss_pos}
\end{align}
Here, $\mathbbm{1}_{(\cdot)}$ denotes the indicator function, and $P_M$ is the number of positive pairs in $\mathcal{D}$. As demonstrated in Section \ref{sec:experiments}, optimizing the above loss not only helps implicitly encode categorical semantic structure into representations, but also promotes better pairwise semantic reasoning capability, though no pairwise supervision except the true entailment labels are present to the model.

\subsection{Learning from Hard Negatives}
\label{sec:hargneg_sample}
Notice that Eq (\ref{eq:pairsupcon_pos}) can be rewritten as
\begin{align}
    &\ell^i_{\text{ID}} = \nonumber \\
    &\log\left(1 + \sum_{j \neq i, i'}\exp\left[\frac{\text{s}(z_{i}, z_j)- \text{s}(z_i, z_{i'})}{\tau}\right]\right)\;. \nonumber 
\end{align}
It can be interpreted as extending the vanilla triplet loss by treating the other $2M$--2 samples within the same minibatch as negatives.  However, the negatives are uniformly sampled from the training data, regardless of how informative they are. Ideally, we want to focus on hard negatives that are from different semantic groups but are mapped close to the anchor, \ie $z_i$, in the representation space. Although the categorical level supervision is not available in NLI, we approximate the importance of the negative samples via the following,
\begin{align}
\label{eq:weighted_instdisc}
    &\ell^i_{\text{wID}}= \\
    & \log\left(1 + \sum_{j \neq i, i'}\exp\left[\frac{\alpha_j\text{s}(z_{i}, z_j)- \text{s}(z_i, z_{i'})}{\tau}\right]\right)\;. \nonumber
\end{align}
Here $\alpha_j =  \frac{\exp(\text{s}(z_i, z_{j})/\tau)}{\frac{1}{2M-2}\sum_{k \neq i, i'} \exp(\text{s}(z_{i}, z_k)/\tau)}$, which can be interpreted as the relative importance of $z_j$ among all $2M$-2 negatives of anchor $z_{i}$. The assumption is that hard negatives are more likely to be those that are located close to the anchor in the representation space. Although there might exists false negatives, \ie those located close to the anchor $z_i$ but are from the same category, the probability is low as long as the underlying number of categories of the training data is not too small and each minibatch $\mathcal{D}$ is uniformly sampled. 


\subsection{Entailment and Contradiction Reasoning}
The instance discrimination loss mainly focuses on separating each positive pair apart from the others, whereas there is no explicit force in discriminating contradiction and entailment.  To this end, we jointly optimize a pairwise entailment and contradiction reasoning objective.  We adopt the softmax-based cross-entropy to form the pairwise classification objective. Let $u_i=\psi(x_i)$ denote the representation of sentence $x_i$, then for each labeled sentence pair $(x_i, x_{i'}, y_i)$ we minimize the following,  
\begin{align}
    \ell_{\text{C}}^i = \text{CE}\left(f(u_i, u_{i'}, |u_i-u_{i'}|), y_i\right)\;.
    \label{eq:pair_classify}
\end{align}
Here $f$ denotes the linear classification head in Figure \ref{fig:motivation_validation}, and CE is the cross-entropy loss.  Different from \citet{reimers2019sentence}, we exclude the neural pairs from the original training set and focus on the binary classification of semantic entailment and contradiction only. Our motivation is that the concept semantic neural can be well captured by the instance discrimination loss. Therefore, we drop the neural pairs from the training data to reduce the functional redundancy of the two losses in PairSupCon and improve the learning efficiency as well. 

\paragraph{Overall loss}{In summary, our overall loss is
\begin{align}
    \mathcal{L} = \sum_{i=1}^{M} \ell_{\text{C}}^i +  \beta \mathbbm{1}_{y_i=1} \cdot \left(\ell_{\text{wID}}^{i} + \ell_{\text{wID}}^{i'}\right)\;,
\end{align}
where $\ell_{\text{C}}^i$ and $\ell_{\text{wID}}^{i}, \ell_{\text{wID}}^{i'}$ are defined in Equations (\ref{eq:pair_classify}) and (\ref{eq:weighted_instdisc}), respectively. In the above equation, $\beta$ balances between the capability of pairwise semantic entailment and contradiction reasoning, and the capability of the high-level categorical semantic structure encoding. We dive deep into the trade-off between these two aspects by evaluating PairSupCon with different $\beta$ values in Section \ref{subsec:ablation}, and show how different $\beta$ values can benefit different downstream tasks. Unless otherwise specified, we set $\beta=1$ in this paper for the purpose of providing effective representations to various downstream tasks instead of being tailored to any specific ones.

}

\begin{table*}[htbp]
  \begin{center}
    \begin{tabular}{lccccccccc}
      & \textbf{AG}& \textbf{SS}&\textbf{SO}& \textbf{Bio}&\textbf{Tweet}& \textbf{G-TS}&\textbf{G-S}& \textbf{G-T}& \textbf{Avg.} \\
      \cline{2-10}
      BERT$_{\text{distil} \cdot \text{base}}$ &\textbf{85.8}&	68.4&	20.7&	32.0&	47.5&	63.2&	56.8&	50.0&	53.1\\
      SBERT$^{\diamondsuit}_{\text{distil} \cdot \text{base}}$&61.1&	54.9&	32.2&	32.6&	45.9&	56.7&	50.7&	48.2&	47.8 \\
      SimCSE$^{\spadesuit}$&-&-&-&-&-&-&-&-&- \\ 
       \hline 
      \textbf{PairSupCon}$_{\text{distil} \cdot \text{base}}$ &82.5&	\textbf{71.5}&	\textbf{63.6}&	\textbf{39.0}&	\textbf{54.2}&	\textbf{69.1}&	\textbf{63.6}&	\textbf{59.7}&	\textbf{62.9} 
      \\ \\
       BERT$_{\text{base}}$ &79.7&	64.0&	21.8&	32.3&	45.1&	61.0&	55.8&	46.7&	50.8 \\
       SBERT$^{\diamondsuit}_{\text{base}}$&65.8&	62.5&	29.6&	31.9&	45.6&	56.5&	50.2&	49.3&	48.9 \\
      SimCSE$^{\spadesuit}$& 82.4&	67.3&	49.0&	37.5&	56.1&	68.1&	63.2&	59.6&	60.4
\\
       \hline 
      \textbf{PairSupCon}$_{\text{base}}$ &\textbf{83.0}&	\textbf{73.6}&	\textbf{63.8}&	\textbf{38.8}&	\textbf{55.7}&	\textbf{69.4}&	\textbf{64.5}&	\textbf{60.4}&	\textbf{63.7}
      \\ \\
      
      BERT$_{\text{large}}$ &83.1&	66.5&	26.4&	31.4&	44.3&	62.3&	55.8&	46.4&	52.0 \\
       SBERT$^{\diamondsuit}_{\text{large}}$ &66.6&	63.7&	37.7&	34.6&	47.7&	59.1&	53.8&	49.5&	51.6 \\
      SimCSE$^{\spadesuit}$& 82.7&	66.3&	49.0&	40.5&	\textbf{57.9}&	68.1&	62.4&	60.8&	61.0
 \\
      \hline 
      \textbf{PairSupCon}$_{\text{large}}$ &\textbf{84.2}&	\textbf{75.7}&	\textbf{63.9}&	\textbf{41.8}&	55.8&	\textbf{70.4}&	\textbf{64.8}&	\textbf{61.3}&	\textbf{64.7} \\
    \end{tabular}
    \caption{Clustering accuracy reported on eight shorttext clustering datasets. The results are averaged over 10 clustering runs using KMeans with independent seeds. ${\diamondsuit}$ and $\spadesuit$: results evaluated on the checkpoints provided by \citet{reimers-2019-sentence-bert} and \citet{gao2021simcse}, respectively.
    }
    \label{tab:Cluster_Eval}
  \end{center}
\end{table*}

\section{Experiments}
\label{sec:experiments}

\paragraph{Baselines}{In this section, we mainly investigate the effective strategies of leveraging the labeled NLI data to enhance the sentence representations of the pre-trained language models (PLMs). We compare PairSupCon against the vanilla BERT \citep{devlin2018bert, sanh2019distilbert} models and the previous state-of-the-art approach, SBERT \citep{reimers-2019-sentence-bert}. We noticed a concurrent work SimCSE \citep{gao2021simcse} when we were preparing this submission. Although SimCSE also leverages the instance discrimination learning to improve the sentence embeddings, it shares a different motivation and focus than ours. We compare PairSupCon against SimCSE to show how different instance discrimination based approaches enhance the sentence representations differently, while our contribution is claimed over the previous SOTA models. Please refer to Appendix \ref{appendix:implementation} for the details of our implementation. 
}

\begin{table}[!htbp]
  \begin{center}
  {\scriptsize
    {\begin{tabular}{l|ccccc}
    \hline
    Dataset & $N$ & $W$ & $C$ & L &S  \\
    \hline
    AgNews (AG) &8K &23 &4 & 2K  &2K \\
    StackOverflow (SO) & 20K &8 & 20 &1K &1K   \\
    Biomedical (Bio)& 20K &13 &  20 &1K &1K \\
    SearchSnippets (SS)& 12K &18 & 8 & 2.66K & .37K  \\
    GooglenewsTS (G-TS)& 11K & 28 & 152 & 430 & 3   \\
    GooglenewsS (G-S)& 11K & 22 & 152 & 430 & 3  \\
    GooglenewsT (G-T)& 11K & 6 & 152 & 430 & 3  \\
    Tweet (Tweet)& 5K & 8 & 89 & 249& 1 \\
      \hline
    \end{tabular}}}
    \caption{Statistics of eight short text clustering datasets. $N$: number of text examples; $W$: average number of words contained in each text example; $C$ number of clusters; L: the size of the largest cluster; and $S$: the size of the smallest cluster. }
    \label{tab:cluster_datastats}
  \end{center}
\end{table}

\subsection{Clustering}
\label{subsec:clustering}
\paragraph{Motivation}{Existing work mainly focuses on the semantic similarity (a.k.a STS) related tasks. We argue that an equally important aspect of sentence representation evaluation -- the capability of encoding the high-level categorical structure into the representations, has so far been neglected. Desirably, a model should map the instances from the same category close together in the representation space while mapping those from different categories farther apart. This expectation aligns well with the underlying assumption of clustering and is consistent with how human categorizes data. We evaluate the capabilities of categorical concept embedding using K-Means \citep{macqueen1967some,lloyd1982least}, given its simplicity and the fact that the algorithm itself manifests the above expectation.

We consider eight benchmark datasets for short text clustering. As indicated in Table \ref{tab:cluster_datastats}\footnote{Please refer to  Appendix \ref{appendix:cluster_data_stats} for more details.}, the datasets present the desired diversities of both the size of each cluster and the number of clusters of each dataset. Furthermore, each text instance consists of 6 to 28 words when averaged within each dataset, which well covers the spectrum of NLI where each instance has 12 words on average. Therefore, we believe the proposed datasets can provide an informative evaluation on whether an embedding model is capable of capturing the high-level categorical concept.
}

\begin{table*}[htbp]
  \begin{center}
    \begin{tabular}{lcccccccc}
      & \textbf{STS12}& \textbf{STS13}&\textbf{STS14}& \textbf{STS15}&\textbf{STS16}& \textbf{SICK-R}&\textbf{STS-B}& \textbf{Avg.} \\
      \cline{2-9}
      BERT$_{\text{distil} \cdot \text{base}}$ &43.4&	64.9&	54.1&	66.6&	68.5&	63.5&	57.2&	59.8 \\
      SBERT$^{\diamondsuit}_{\text{distil} \cdot \text{base}}$&69.8&	72.2&	70.7&	79.9&	75.4&	74.5&	78.4&	74.4 \\
      SimCSE$^{\spadesuit}$&-&-&-&-&-&-&-&- \\ 
      \hline 
      \textbf{PairSupCon}$_{\text{distil} \cdot \text{base}}$ &\textbf{73.6}&	\textbf{84.1}&	\textbf{78.3}&	\textbf{84.4}&	\textbf{81.8}&	\textbf{77.8}&	\textbf{82.3}&	\textbf{80.3}
      \\ \\
      BERT$_{\text{base}}$ &30.9&	59.9&	47.7&	60.3&	63.7&	58.2&	47.3&	52.6 \\
      SBERT$^{\diamondsuit}_{\text{base}}$ &71.0&	76.5&	73.2&	79.1&	74.3&	72.9&	77.0&	74.9 \\
      SimCSE$^{\spadesuit}$& \textbf{75.3}& \textbf{84.7}& \textbf{80.2}& \textbf{85.4}& 80.8& \textbf{80.4}& \textbf{84.3}& \textbf{81.6}\\
      \hline 
      \textbf{PairSupCon}$_{\text{base}}$ &74.3&	84.4&	79.0&	84.8&	\textbf{81.4}&	79.0&	82.6&	80.8
      \\ \\
     BERT$_{\text{large}}$ &27.7&	55.8&	44.5&	51.7&	61.9&	53.9&	47.0&	48.9 \\
     SBERT$^{\diamondsuit}_{\text{large}}$ &72.3&	78.5&	74.9&	81.0&	76.3&	73.8&	79.2&	76.5 \\
      SimCSE$^{\spadesuit}$&\textbf{75.8}& \textbf{86.3}& \textbf{80.4}& \textbf{86.1}& 80.9& \textbf{81.1}& \textbf{84.9}&   \textbf{82.2} \\
      \hline 
      \textbf{PairSupCon}$_{\text{large}}$ &74.2&	85.8&	79.5&	85.5&	\textbf{81.7}&	80.4&	83.7&	81.5 \\
    \end{tabular}
    \caption{Spearman rank correlation between the cosine similarity of sentence representations and the ground truth labels on seven Semantic Textual Similarity (STS) tasks. $\diamondsuit$ and $\spadesuit$: results evaluated on the checkpoints provided by \citet{reimers-2019-sentence-bert} and \citet{gao2021simcse}, respectively.
    }
    \label{tab:STS_compare}
  \end{center}
\end{table*}

\paragraph{Evaluation Results}{The evaluation results are summarized in Table \ref{tab:Cluster_Eval}. We run K-Means with the scikit-learn package \citep{scikit-learn} on the representations provided by each model and report the clustering accuracy  \footnote{We computed the clustering accuracy using the Hungarian algorithm \citep{munkres1957algorithms}.} averaged over 10 independent runs. \footnote{We randomly select 10 independent seeds and fix them for all evaluations} As Table \ref{tab:Cluster_Eval} indicates, in comparison with the vanilla BERT models, SBERT results in degenerated embedding of the categorical semantic structure by simply optimizing the pairwise siamese loss.  One possible reason is that  SBERT uses a large learning rate (2e-05) to optimize the transformer, which can cause catastrophic forgetting of the knowledge acquired in the original BERT models. We find using a smaller learning rate for the backbone can consistently improve the performance of SBERT (see the performance of BERT-base with "Classificaiton" in Table \ref{tab:PairSupCon_ablation}).

Nevertheless, PairSupCon leads to an averaged improvement of $10.8\%$ to $15.2\%$ over SBERT, which validates our motivation in leveraging the implicit grouping effect of the instance discrimination learning to better encode the high-level semantic concepts into representations. Moreover, PairSupCon also attains better performance than SimCSE, and we suspect this is because PairSupCon better leverages the training data. Specifically, PairSupCon aims to discriminate an positive (entailment) sentence pair apart from all other sentences, no matter they are premises or hypotheses. In contrast, SimCSE only separates a premise from the other premises through their entailment and contradiction hypotheses, while there is no explicit instance discrimination force within the premises or the hypotheses alone. Considering the statistic data difference \citep{williams2017broad} between premises and hypotheses, PairSupCon can potentially better capture categorical semantic concepts by leveraging additional intrinsic semantic properties of the premises or the hypotheses that are undiscovered by SimCSE.
}

\begin{figure*}[htbp]
    \centering
    \begin{subfigure}{0.92\textwidth}
        \centering
         \includegraphics[scale=0.36]{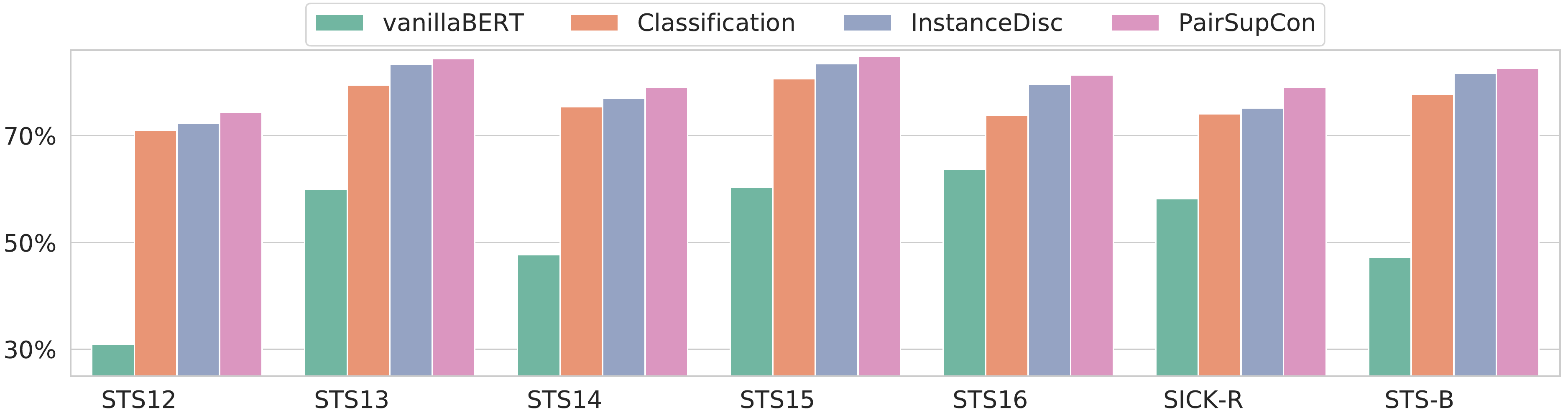}
        \label{fig:stshist_ablation}
    \end{subfigure}
    \vspace{0.1cm}
    
    \begin{subfigure}{0.9\textwidth}
        \centering
        \includegraphics[scale=0.33]{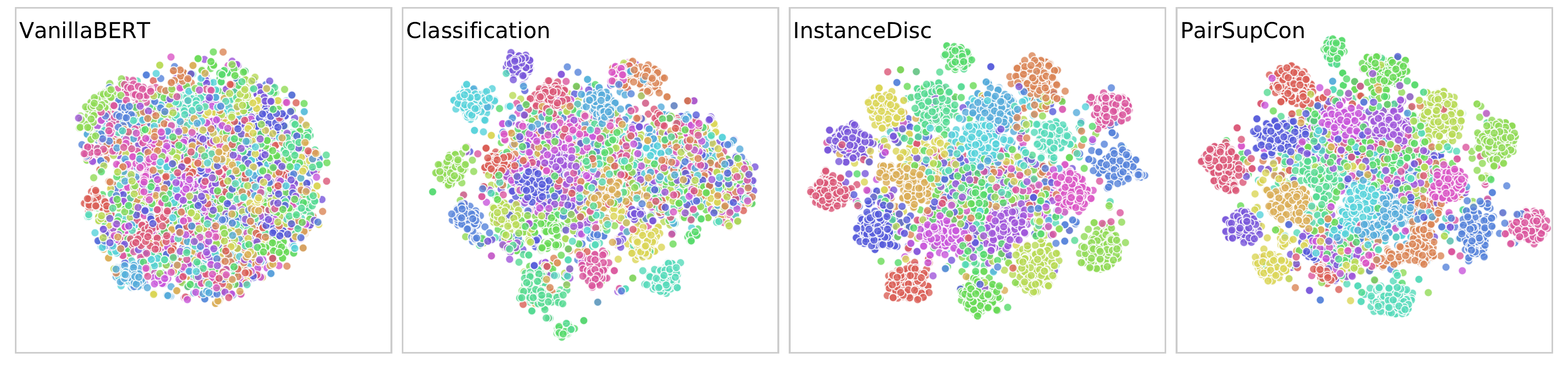}
        \label{fig:tsne_stackoverflow_ablation}
    \end{subfigure}
    \caption{Ablation study of PairSupCon. \textbf{First Row}: Spearman correlation evaluated on the STS tasks. \textbf{Second Row}: visualization of the StackOverflow \citep{xu2017self} representations using t-SNE \citep{maaten2008visualizing}.}
    \label{fig:motivation_ablation}
\end{figure*}

\subsection{Semantic Textual Similarity}
\label{subsec:sts}
Next, we asses the performance of PairSupCon on seven STS tasks, namely  STS 2012-2016 \citep{agirre2012semeval,agirre2013sem,agirre2014semeval,agirre2015semeval,agirre2016semeval}, STS Benchmark \citep{cer2017semeval}, and SICK-Relatedness  \citep{marelli2014sick}.  These datasets include pairs of sentences with a fine-grained gold semantic similarity score ranges from 0 to 5. To enable a fair comparison, we follow the setup in SBERT \citep{reimers-2019-sentence-bert}, and compute the Spearman correlation\footnote{Same as SBERT \citep{reimers-2019-sentence-bert} and SimCSE \citep{gao2021simcse}, we concatenate all the topics and report the overall Spearman’s correlation.} between cosine similarity of sentence embeddings and the ground truth similarity scores of each pair. 

The evaluation results are reported in Table \ref{tab:STS_compare}. PairSupCon substantially outperforms both the vanilla BERT and SBERT models. This validates our assumption that, by implicitly encoding the high-level categorical structure into the representations, PairSupCon promotes better convergence of the low-level semantic entailment reasoning objective. This assumption is consistent with the top-down categorization behavior of humans. Although SimCSE leverages STS-Benchmark as the development set while PairSupCon is fully blind to the downstream tasks\footnote{We also evaluate PairSupCon with STS-Benchmark as the development set in Appendix \ref{appendix:sts_dev}.}, we hypothesize the performance gain of SimCSE on STS is mainly contributed by explicitly merging the entailment and contradiction separation into the instance discrimination loss. On the other hand, as we discussed in Section \ref{subsec:clustering}, PairSupCon achieves more obvious performance gain on the clustering tasks through a bidirectional instance discrimination loss. Therefore, developing a better instance discrimination based sentence representation learning objective by incorporating the strengths of both SimCSE and PairSupCon could be a promising direction. 
\begin{table}[htbp]
  \begin{center}
    \begin{tabular}{@{\extracolsep{8pt}}ccc@{}}
      \textbf{Loss Func.} & \multicolumn{2}{c}{\textbf{Downstream}} \\
      \cline{2-3} 
      & \textbf{STS}  & \textbf{Clustering}\\
      Classification & 76.0& 52.9 \\
      InstanceDisc &79.0 & \textbf{65.5} \\
      \hline 
      \hline 
      \textbf{PairSupCon}& \\
      $\beta=0.5$ & \textbf{81.0} & 61.6 \\
      $\underline{\beta=1}$ & 80.8 & 63.7 \\
      $\beta=2$ & 80.4 & 64.6 \\
      $\beta=4$ & 80.0 & 65.0 \\
      $\beta=4$ & 79.5 & 65.4 \\
      \hline 
    \end{tabular}
    \caption{Ablation study of PiarSupCon on BERT-base. We underline the setting used for all our experiments.}
    \label{tab:PairSupCon_ablation}
  \end{center}
\end{table}

\begin{table}[htbp]
  \begin{center}
    \begin{tabular}{@{\extracolsep{5pt}}lcc@{}}
      \textbf{Loss Func.} & \multicolumn{2}{c}{\textbf{Downstream}} \\
      \cline{2-3} 
      & \textbf{STS}  & \textbf{Clustering}\\
      \textbf{InstanceDisc} & \\
      w/o HardNegSample &  78.4 & 64.3\\
      w/ HardNegSample & \textbf{79.0} & \textbf{65.5} \\
      \hline 
      \hline 
      \textbf{PairSupCon}& \\
      w/o HardNegSample & 80.2 & 62.1\\
      w/ HardNegSample &  \textbf{80.8} & \textbf{63.7}\\
      \hline 
    \end{tabular}
    \caption{Ablation of hard negative sampling. We choose BERT-base as backbone.}
    \label{tab:importance}
  \end{center}
\end{table}

\subsection{Ablation Study}
\label{subsec:ablation}
We run extensive ablations to better understand what enables the good performance of PairSupCon. For notational convenience, we name the pairwise semantic relation classification objective in PairSupCon as \textbf{Classification}, and the instance discrimination objective as \textbf{InstanceDisc}. 
\begin{table*}[htbp]
  \begin{center}
  {\small{
    \begin{tabular}{lccccccc}
      & \textbf{MR}& \textbf{CR}&\textbf{SUBJ}& \textbf{MPQA}&\textbf{SST}& \textbf{TREC}&\textbf{MRPC} \\
      \cline{2-8}
      BERT$_{\text{distil} \cdot \text{base}}$ &61.4(4.0)	&69.0(2.6)	&\textbf{86.2}(1.8)	&67.4(3.8)	&67.3(4.2)	&\textbf{59.2}(6.3)	&59.8(6.9) \\
      SBERT$^{\diamondsuit}_{\text{distil} \cdot \text{base}}$&\textbf{69.8}(1.9) &\textbf{83.3}(1.9)	&80.4(	2.8)	&81.9(1.5)	&78.8(1.2)	&39.8(4.2)	&\textbf{60.4}(3.4) \\
      SimCSE$^{\spadesuit}$&-&-&-&-&-&-&- \\ 
      \hline 
      \textbf{PairSupCon}$_{\text{distil} \cdot \text{base}}$ &65.6	(6.4)	&80.5(1.5)	&82.7	(3.3)	&\textbf{83.8}(1.1)	&\textbf{79.9}(1.6)	&52.5(3.2) &59.7(2.0)
      \\ \\
      BERT$_{\text{base}}$ &61.8(4.3)	&68.2(3.1)	&\textbf{87.4}(1.6)	&63.6(5.2)	&68.0(1.7)	&\textbf{58.4}(2.1)	&56.5(12.1)\\
      SBERT$^{\diamondsuit}_{\text{base}}$ &\textbf{72.9}(1.7)	&81.2(4.9)	&81.9(1.1)	&80.7(2.4)	&78.0	(6.1)	&45.8(1.9)	&60.0(3.9)\\
      SimCSE$^{\spadesuit}$&71.8(1.8)	&\textbf{82.9}(1.5)	&85.9(2.0)	&79.0(4.2)	&80.0(3.7)	&54.5(4.0)	&57.8(5.1) \\
      \hline 
      \textbf{PairSupCon}$_{\text{base}}$ &72.8(2.0)	&81.8(2.6)	&85.7(1.9)	&\textbf{82.7}(1.2)	&\textbf{80.2}(3.2)	&49.3(3.9) &\textbf{61.4}(4.4)
      \\ \\
     BERT$_{\text{large}}$ &64.4(4.4)	&70.9(2.9)	&\textbf{86.6}(3.0)	&62.9(4.6)	&73.1(3.0)	&\textbf{57.2}(2.9) &\textbf{63.4}(9.2) \\
     SBERT$^{\diamondsuit}_{\text{large}}$ &\textbf{78.5}(0.9)	&\textbf{85.1}(6.3)	&79.8(3.6)	&83.3(1.6)	&82.5(4.8)	&43.5(4.9)	&58.8(4.7) \\
      SimCSE$^{\spadesuit}$&75.4(2.3)	&84.5(3.0)	&85.5(2.8)	&84.0(1.9)	&\textbf{84.9}(2.2)	&52.4(3.4)	&59.6(2.4) \\
      \hline 
      \textbf{PairSupCon}$_{\text{large}}$ &74.8(2.6)	&84.8(4.1)	&84.6(2.3)	&\textbf{85.2}(1.3)	&83.7(1.7)	&50.4(5.7)	&62.6(5.3) \\
    \end{tabular}}}
    \caption{Few-shot learning evaluation on SentEval. For each task, we randomly sample 16 labeled instances per class and report the mean
(standard deviation) performance over 5 different training sets. $\diamondsuit$ and $\spadesuit$: results evaluated on the checkpoints provided by \citet{reimers-2019-sentence-bert} and \citet{gao2021simcse}, respectively. 
    }
    \label{tab:senteval_fsl_compare}
  \end{center}
\end{table*}
\paragraph{PairSupCon versus. Its Components}{In Figure \ref{fig:motivation_ablation}, we compare PairSupCon against its two components, namely Classification and InstanceDisc. As it shows, InstanceDisc itself outperforms Classification on both STS and categorical concept encoding. The result matches our expectation that contrasting each positive pair against multiple negatives, despite obtained through unsupervised sampling, yields better performance than simply learning from each individual pair. By jointly optimizing both objectives, PairSupCon leverages the implicit grouping effect of InstanceDisc to encode the high-level categorical structure into representations while, simultaneously complementing InstanceDisc with more fine-grained semantic concept reasoning capability via Classification.

Table \ref{tab:PairSupCon_ablation} indicates a trade-off between the high-level semantic structure encoding and the low-level pairwise entailment and contradiction reasoning capability of PairSupCon. Focusing more on the pairwise classification objective, \ie using smaller $\beta$ values, can hurt the embeddings of the high-level semantic structure. This result is not surprising, especially considering that sentences forming a contradiction pair are not necessarily belong to different semantic groups. On the other hand, InstanceDisc only focuses on separating each positive pair from all other samples within the same minibatch, and an explicit force that discriminates semantic entailment from contradiction is necessary for PairSupCon to achieve competitive performance on the more fine-grained pairwise similarity reasoning on STS. As indicated in Table \ref{tab:PairSupCon_ablation}, we can tune the $\beta$ values to attain effective representations for specific downstream tasks according to their semantic granularity focuses. We set $\beta=1$ for all our experiments with the goal to provide effective universal sentence representations to different downstream tasks.


}

\paragraph{Hard Negative Sampling Helps}{
In Table \ref{tab:importance}, we compare both PairSupCon and InstanceDisc against their counterparts where the negatives in the instance discrimination loss are uniformly sampled from data. As it shows, the hard negative sampling approach proposed in Section \ref{sec:hargneg_sample} leads to improved performance on both STS and clustering tasks. We associate this performance boost with our assumption that hard negatives are likely located close to the anchor. A properly designed distance-based sampling approach can drive the model to better focus on hard negative separation and hence lead to better performance. 

On the other hand, hard negative sampling without any supervision is a very challenging problem, especially considering that samples within the local region of an anchor are also likely from the same semantic group as the anchor. As a consequence, a solely distance-based sampling approach can induce certain false negatives and hurt the performance. To tackle this issue, leveraging proper structure assumption or domain-specific knowledge could be potential directions, which we leave as future work.   
}

\subsection{Transfer Learning}
\label{subsec:transfer_learning}
In order to provide a fair and comprehensive comparison with the existing work, we also evaluate PairSupCon on the following seven transferring tasks: MR \citep{pang2005seeing}, CR \citep{hu2004mining}, SUBJ \citep{pang2004sentimental}, MPQA \citep{wiebe2005annotating}, SST \citep{socher2013recursive}, TREC \citep{li2002learning}, and MRPC \citep{dolan2004unsupervised}. We follow the widely used evaluation protocol, where a logistic regression classifier is trained on top of the frozen representations, and the testing accuracy is used as a measure of the representation quality. We adopt the default configurations of the SentEval \citep{conneau2018senteval} toolkit and report the evaluation results in Table \ref{tab:senteval_compare_full_supervision} in Appendix \ref{appendix:transfer_learning}. 
As we can see, the performance gap between different methods are relatively small. 

We suspect the reason is that the transfer learning tasks do not present enough complexities to fully uncover the performance gap between different approaches, especially considering that most tasks are binary classification with a large amount of labeled training examples. To further examine our hypothesis, we extend the evaluation to the setting of few-shot learning, where we uniformly sample 16 labeled instances per class for each task.  We report the mean
and standard deviation of the evaluation performance over 5 different sample sets in Table \ref{tab:senteval_fsl_compare}.  Although we observe more obvious performance gap on each specific task, there is no consistent performance gap between different approaches when evaluated across different tasks. Therefore, to better evaluate the transfer learning performance of sentence representations, more complex and diverse datasets are required.  

\section{Discussion and Conclusion}
In this paper, we present a simple framework for universal sentence representation learning. We leverage the implicit grouping effect of instance discrimination learning to better encoding the high-level semantic structure of data into representations while, simultaneously promoting better convergence of the lower-level semantic entailment and contradiction reasoning objective. We substantially advance the previous state-of-the-art results when evaluated on various downstream tasks that involve understanding semantic concepts at different granularities. 

We carefully study the key components of our model and pinpoint the performance gain contributed by each of them.  We observe encouraging performance improvement by using the proposed hard negative sampling strategy. On the other hand, hard negative sampling without any supervision is an crucial, yet significantly challenging problem that should motivate further explorations. Possible directions include making proper structure assumption or leveraging domain-specific knowledge. The substantial performance gain attained by our model also suggests developing explicit grouping objectives could be another direction worth investigation. 

\bibliography{anthology,emnlp2020}

\begin{thebibliography}{54}
\expandafter\ifx\csname natexlab\endcsname\relax\def\natexlab#1{#1}\fi

\bibitem[{Agirre et~al.(2015)Agirre, Banea, Cardie, Cer, Diab, Gonzalez-Agirre,
  Guo, Lopez-Gazpio, Maritxalar, Mihalcea et~al.}]{agirre2015semeval}
Eneko Agirre, Carmen Banea, Claire Cardie, Daniel Cer, Mona Diab, Aitor
  Gonzalez-Agirre, Weiwei Guo, Inigo Lopez-Gazpio, Montse Maritxalar, Rada
  Mihalcea, et~al. 2015.
\newblock Semeval-2015 task 2: Semantic textual similarity, english, spanish
  and pilot on interpretability.
\newblock In \emph{Proceedings of the 9th international workshop on semantic
  evaluation (SemEval 2015)}, pages 252--263.

\bibitem[{Agirre et~al.(2014)Agirre, Banea, Cardie, Cer, Diab, Gonzalez-Agirre,
  Guo, Mihalcea, Rigau, and Wiebe}]{agirre2014semeval}
Eneko Agirre, Carmen Banea, Claire Cardie, Daniel Cer, Mona Diab, Aitor
  Gonzalez-Agirre, Weiwei Guo, Rada Mihalcea, German Rigau, and Janyce Wiebe.
  2014.
\newblock Semeval-2014 task 10: Multilingual semantic textual similarity.
\newblock In \emph{Proceedings of the 8th international workshop on semantic
  evaluation (SemEval 2014)}, pages 81--91.

\bibitem[{Agirre et~al.(2016)Agirre, Banea, Cer, Diab, Gonzalez~Agirre,
  Mihalcea, Rigau~Claramunt, and Wiebe}]{agirre2016semeval}
Eneko Agirre, Carmen Banea, Daniel Cer, Mona Diab, Aitor Gonzalez~Agirre, Rada
  Mihalcea, German Rigau~Claramunt, and Janyce Wiebe. 2016.
\newblock Semeval-2016 task 1: Semantic textual similarity, monolingual and
  cross-lingual evaluation.
\newblock In \emph{SemEval-2016. 10th International Workshop on Semantic
  Evaluation; 2016 Jun 16-17; San Diego, CA. Stroudsburg (PA): ACL; 2016. p.
  497-511.} ACL (Association for Computational Linguistics).

\bibitem[{Agirre et~al.(2012)Agirre, Cer, Diab, and
  Gonzalez-Agirre}]{agirre2012semeval}
Eneko Agirre, Daniel Cer, Mona Diab, and Aitor Gonzalez-Agirre. 2012.
\newblock Semeval-2012 task 6: A pilot on semantic textual similarity.
\newblock In \emph{* SEM 2012: The First Joint Conference on Lexical and
  Computational Semantics--Volume 1: Proceedings of the main conference and the
  shared task, and Volume 2: Proceedings of the Sixth International Workshop on
  Semantic Evaluation (SemEval 2012)}, pages 385--393.

\bibitem[{Agirre et~al.(2013)Agirre, Cer, Diab, Gonzalez-Agirre, and
  Guo}]{agirre2013sem}
Eneko Agirre, Daniel Cer, Mona Diab, Aitor Gonzalez-Agirre, and Weiwei Guo.
  2013.
\newblock * sem 2013 shared task: Semantic textual similarity.
\newblock In \emph{Second joint conference on lexical and computational
  semantics (* SEM), volume 1: proceedings of the Main conference and the
  shared task: semantic textual similarity}, pages 32--43.

\bibitem[{Bachman et~al.(2019)Bachman, Hjelm, and
  Buchwalter}]{bachman2019learning}
Philip Bachman, R~Devon Hjelm, and William Buchwalter. 2019.
\newblock Learning representations by maximizing mutual information across
  views.
\newblock In \emph{Advances in Neural Information Processing Systems}, pages
  15535--15545.

\bibitem[{Bowman et~al.(2015)Bowman, Angeli, Potts, and
  Manning}]{bowman2015large}
Samuel~R Bowman, Gabor Angeli, Christopher Potts, and Christopher~D Manning.
  2015.
\newblock A large annotated corpus for learning natural language inference.
\newblock \emph{arXiv preprint arXiv:1508.05326}.

\bibitem[{Cer et~al.(2017)Cer, Diab, Agirre, Lopez-Gazpio, and
  Specia}]{cer2017semeval}
Daniel Cer, Mona Diab, Eneko Agirre, Inigo Lopez-Gazpio, and Lucia Specia.
  2017.
\newblock Semeval-2017 task 1: Semantic textual similarity-multilingual and
  cross-lingual focused evaluation.
\newblock \emph{arXiv preprint arXiv:1708.00055}.

\bibitem[{Cer et~al.(2018)Cer, Yang, Kong, Hua, Limtiaco, John, Constant,
  Guajardo-C{\'e}spedes, Yuan, Tar et~al.}]{cer2018universal}
Daniel Cer, Yinfei Yang, Sheng-yi Kong, Nan Hua, Nicole Limtiaco, Rhomni~St
  John, Noah Constant, Mario Guajardo-C{\'e}spedes, Steve Yuan, Chris Tar,
  et~al. 2018.
\newblock Universal sentence encoder.
\newblock \emph{arXiv preprint arXiv:1803.11175}.

\bibitem[{Chen et~al.(2020)Chen, Kornblith, Norouzi, and
  Hinton}]{chen2020simple}
Ting Chen, Simon Kornblith, Mohammad Norouzi, and Geoffrey Hinton. 2020.
\newblock A simple framework for contrastive learning of visual
  representations.
\newblock \emph{arXiv preprint arXiv:2002.05709}.

\bibitem[{Conneau and Kiela(2018)}]{conneau2018senteval}
Alexis Conneau and Douwe Kiela. 2018.
\newblock Senteval: An evaluation toolkit for universal sentence
  representations.
\newblock \emph{arXiv preprint arXiv:1803.05449}.

\bibitem[{Conneau et~al.(2017)Conneau, Kiela, Schwenk, Barrault, and
  Bordes}]{conneau2017supervised}
Alexis Conneau, Douwe Kiela, Holger Schwenk, Loic Barrault, and Antoine Bordes.
  2017.
\newblock Supervised learning of universal sentence representations from
  natural language inference data.
\newblock \emph{arXiv preprint arXiv:1705.02364}.

\bibitem[{Devlin et~al.(2018)Devlin, Chang, Lee, and
  Toutanova}]{devlin2018bert}
Jacob Devlin, Ming-Wei Chang, Kenton Lee, and Kristina Toutanova. 2018.
\newblock Bert: Pre-training of deep bidirectional transformers for language
  understanding.
\newblock \emph{arXiv preprint arXiv:1810.04805}.

\bibitem[{Dolan et~al.(2004)Dolan, Quirk, Brockett, and
  Dolan}]{dolan2004unsupervised}
William Dolan, Chris Quirk, Chris Brockett, and Bill Dolan. 2004.
\newblock Unsupervised construction of large paraphrase corpora: Exploiting
  massively parallel news sources.

\bibitem[{Gao et~al.(2021)Gao, Yao, and Chen}]{gao2021simcse}
Tianyu Gao, Xingcheng Yao, and Danqi Chen. 2021.
\newblock Simcse: Simple contrastive learning of sentence embeddings.
\newblock \emph{arXiv preprint arXiv:2104.08821}.

\bibitem[{Giorgi et~al.(2020)Giorgi, Nitski, Bader, and
  Wang}]{giorgi2020declutr}
John~M Giorgi, Osvald Nitski, Gary~D Bader, and Bo~Wang. 2020.
\newblock Declutr: Deep contrastive learning for unsupervised textual
  representations.
\newblock \emph{arXiv preprint arXiv:2006.03659}.

\bibitem[{Hadsell et~al.(2006)Hadsell, Chopra, and
  LeCun}]{hadsell2006dimensionality}
Raia Hadsell, Sumit Chopra, and Yann LeCun. 2006.
\newblock Dimensionality reduction by learning an invariant mapping.
\newblock In \emph{2006 IEEE Computer Society Conference on Computer Vision and
  Pattern Recognition (CVPR'06)}, volume~2, pages 1735--1742. IEEE.

\bibitem[{Harwood et~al.(2017)Harwood, Kumar~BG, Carneiro, Reid, and
  Drummond}]{harwood2017smart}
Ben Harwood, Vijay Kumar~BG, Gustavo Carneiro, Ian Reid, and Tom Drummond.
  2017.
\newblock Smart mining for deep metric learning.
\newblock In \emph{Proceedings of the IEEE International Conference on Computer
  Vision}, pages 2821--2829.

\bibitem[{He et~al.(2020)He, Fan, Wu, Xie, and Girshick}]{he2020momentum}
Kaiming He, Haoqi Fan, Yuxin Wu, Saining Xie, and Ross Girshick. 2020.
\newblock Momentum contrast for unsupervised visual representation learning.
\newblock In \emph{Proceedings of the IEEE/CVF Conference on Computer Vision
  and Pattern Recognition}, pages 9729--9738.

\bibitem[{Hu and Liu(2004)}]{hu2004mining}
Minqing Hu and Bing Liu. 2004.
\newblock Mining and summarizing customer reviews.
\newblock In \emph{Proceedings of the tenth ACM SIGKDD international conference
  on Knowledge discovery and data mining}, pages 168--177.

\bibitem[{Kingma and Ba(2015)}]{KingmaB14}
Diederik~P. Kingma and Jimmy Ba. 2015.
\newblock \href {http://arxiv.org/abs/1412.6980} {Adam: {A} method for
  stochastic optimization}.
\newblock In \emph{3rd International Conference on Learning Representations,
  {ICLR} 2015, San Diego, CA, USA, May 7-9, 2015, Conference Track
  Proceedings}.

\bibitem[{Li and Roth(2002)}]{li2002learning}
Xin Li and Dan Roth. 2002.
\newblock Learning question classifiers.
\newblock In \emph{COLING 2002: The 19th International Conference on
  Computational Linguistics}.

\bibitem[{Lloyd(1982)}]{lloyd1982least}
Stuart Lloyd. 1982.
\newblock Least squares quantization in pcm.
\newblock \emph{IEEE transactions on information theory}, 28(2):129--137.

\bibitem[{Maaten and Hinton(2008)}]{maaten2008visualizing}
Laurens van~der Maaten and Geoffrey Hinton. 2008.
\newblock Visualizing data using t-sne.
\newblock \emph{Journal of machine learning research}, 9(Nov):2579--2605.

\bibitem[{MacQueen et~al.(1967)}]{macqueen1967some}
James MacQueen et~al. 1967.
\newblock Some methods for classification and analysis of multivariate
  observations.
\newblock In \emph{Proceedings of the fifth Berkeley symposium on mathematical
  statistics and probability}, volume~1, pages 281--297. Oakland, CA, USA.

\bibitem[{Marelli et~al.(2014)Marelli, Menini, Baroni, Bentivogli, Bernardi,
  Zamparelli et~al.}]{marelli2014sick}
Marco Marelli, Stefano Menini, Marco Baroni, Luisa Bentivogli, Raffaella
  Bernardi, Roberto Zamparelli, et~al. 2014.
\newblock A sick cure for the evaluation of compositional distributional
  semantic models.
\newblock In \emph{Lrec}, pages 216--223. Reykjavik.

\bibitem[{Meng et~al.(2021)Meng, Xiong, Bajaj, Tiwary, Bennett, Han, and
  Song}]{meng2021coco}
Yu~Meng, Chenyan Xiong, Payal Bajaj, Saurabh Tiwary, Paul Bennett, Jiawei Han,
  and Xia Song. 2021.
\newblock Coco-lm: Correcting and contrasting text sequences for language model
  pretraining.
\newblock \emph{arXiv preprint arXiv:2102.08473}.

\bibitem[{Munkres(1957)}]{munkres1957algorithms}
James Munkres. 1957.
\newblock Algorithms for the assignment and transportation problems.
\newblock \emph{Journal of the society for industrial and applied mathematics},
  5(1):32--38.

\bibitem[{Oh~Song et~al.(2016)Oh~Song, Xiang, Jegelka, and
  Savarese}]{oh2016deep}
Hyun Oh~Song, Yu~Xiang, Stefanie Jegelka, and Silvio Savarese. 2016.
\newblock Deep metric learning via lifted structured feature embedding.
\newblock In \emph{Proceedings of the IEEE conference on computer vision and
  pattern recognition}, pages 4004--4012.

\bibitem[{Oord et~al.(2018)Oord, Li, and Vinyals}]{oord2018representation}
Aaron van~den Oord, Yazhe Li, and Oriol Vinyals. 2018.
\newblock Representation learning with contrastive predictive coding.
\newblock \emph{arXiv preprint arXiv:1807.03748}.

\bibitem[{Pang and Lee(2004)}]{pang2004sentimental}
Bo~Pang and Lillian Lee. 2004.
\newblock A sentimental education: Sentiment analysis using subjectivity
  summarization based on minimum cuts.
\newblock \emph{arXiv preprint cs/0409058}.

\bibitem[{Pang and Lee(2005)}]{pang2005seeing}
Bo~Pang and Lillian Lee. 2005.
\newblock Seeing stars: Exploiting class relationships for sentiment
  categorization with respect to rating scales.
\newblock \emph{arXiv preprint cs/0506075}.

\bibitem[{Pedregosa et~al.(2011)Pedregosa, Varoquaux, Gramfort, Michel,
  Thirion, Grisel, Blondel, Prettenhofer, Weiss, Dubourg, Vanderplas, Passos,
  Cournapeau, Brucher, Perrot, and Duchesnay}]{scikit-learn}
F.~Pedregosa, G.~Varoquaux, A.~Gramfort, V.~Michel, B.~Thirion, O.~Grisel,
  M.~Blondel, P.~Prettenhofer, R.~Weiss, V.~Dubourg, J.~Vanderplas, A.~Passos,
  D.~Cournapeau, M.~Brucher, M.~Perrot, and E.~Duchesnay. 2011.
\newblock Scikit-learn: Machine learning in {P}ython.
\newblock \emph{Journal of Machine Learning Research}, 12:2825--2830.

\bibitem[{Phan et~al.(2008)Phan, Nguyen, and Horiguchi}]{phan2008learning}
Xuan-Hieu Phan, Le-Minh Nguyen, and Susumu Horiguchi. 2008.
\newblock Learning to classify short and sparse text \& web with hidden topics
  from large-scale data collections.
\newblock In \emph{Proceedings of the 17th international conference on World
  Wide Web}, pages 91--100.

\bibitem[{Rakib et~al.(2020)Rakib, Zeh, Jankowska, and
  Milios}]{rakib2020enhancement}
Md~Rashadul~Hasan Rakib, Norbert Zeh, Magdalena Jankowska, and Evangelos
  Milios. 2020.
\newblock Enhancement of short text clustering by iterative classification.
\newblock \emph{arXiv preprint arXiv:2001.11631}.

\bibitem[{Reimers and
  Gurevych(2019{\natexlab{a}})}]{reimers-2019-sentence-bert}
Nils Reimers and Iryna Gurevych. 2019{\natexlab{a}}.
\newblock \href {http://arxiv.org/abs/1908.10084} {Sentence-bert: Sentence
  embeddings using siamese bert-networks}.
\newblock In \emph{Proceedings of the 2019 Conference on Empirical Methods in
  Natural Language Processing}. Association for Computational Linguistics.

\bibitem[{Reimers and Gurevych(2019{\natexlab{b}})}]{reimers2019sentence}
Nils Reimers and Iryna Gurevych. 2019{\natexlab{b}}.
\newblock Sentence-bert: Sentence embeddings using siamese bert-networks.
\newblock \emph{arXiv preprint arXiv:1908.10084}.

\bibitem[{Rethmeier and Augenstein(2021)}]{rethmeier2021primer}
Nils Rethmeier and Isabelle Augenstein. 2021.
\newblock A primer on contrastive pretraining in language processing: Methods,
  lessons learned and perspectives.
\newblock \emph{arXiv preprint arXiv:2102.12982}.

\bibitem[{Sanh et~al.(2019)Sanh, Debut, Chaumond, and
  Wolf}]{sanh2019distilbert}
Victor Sanh, Lysandre Debut, Julien Chaumond, and Thomas Wolf. 2019.
\newblock Distilbert, a distilled version of bert: smaller, faster, cheaper and
  lighter.
\newblock \emph{arXiv preprint arXiv:1910.01108}.

\bibitem[{Schroff et~al.(2015)Schroff, Kalenichenko, and
  Philbin}]{schroff2015facenet}
Florian Schroff, Dmitry Kalenichenko, and James Philbin. 2015.
\newblock Facenet: A unified embedding for face recognition and clustering.
\newblock In \emph{Proceedings of the IEEE conference on computer vision and
  pattern recognition}, pages 815--823.

\bibitem[{Socher et~al.(2013)Socher, Perelygin, Wu, Chuang, Manning, Ng, and
  Potts}]{socher2013recursive}
Richard Socher, Alex Perelygin, Jean Wu, Jason Chuang, Christopher~D Manning,
  Andrew~Y Ng, and Christopher Potts. 2013.
\newblock Recursive deep models for semantic compositionality over a sentiment
  treebank.
\newblock In \emph{Proceedings of the 2013 conference on empirical methods in
  natural language processing}, pages 1631--1642.

\bibitem[{Thakur et~al.(2020)Thakur, Reimers, Daxenberger, and
  Gurevych}]{thakur2020augmented}
Nandan Thakur, Nils Reimers, Johannes Daxenberger, and Iryna Gurevych. 2020.
\newblock Augmented sbert: Data augmentation method for improving bi-encoders
  for pairwise sentence scoring tasks.
\newblock \emph{arXiv preprint arXiv:2010.08240}.

\bibitem[{Wang et~al.(2018)Wang, Singh, Michael, Hill, Levy, and
  Bowman}]{wang2018glue}
Alex Wang, Amanpreet Singh, Julian Michael, Felix Hill, Omer Levy, and Samuel~R
  Bowman. 2018.
\newblock Glue: A multi-task benchmark and analysis platform for natural
  language understanding.
\newblock \emph{arXiv preprint arXiv:1804.07461}.

\bibitem[{Wang et~al.(2014)Wang, Song, Leung, Rosenberg, Wang, Philbin, Chen,
  and Wu}]{wang2014learning}
Jiang Wang, Yang Song, Thomas Leung, Chuck Rosenberg, Jingbin Wang, James
  Philbin, Bo~Chen, and Ying Wu. 2014.
\newblock Learning fine-grained image similarity with deep ranking.
\newblock In \emph{Proceedings of the IEEE conference on computer vision and
  pattern recognition}, pages 1386--1393.

\bibitem[{Weinberger and Saul(2009)}]{weinberger2009distance}
Kilian~Q Weinberger and Lawrence~K Saul. 2009.
\newblock Distance metric learning for large margin nearest neighbor
  classification.
\newblock \emph{Journal of machine learning research}, 10(2).

\bibitem[{Wiebe et~al.(2005)Wiebe, Wilson, and Cardie}]{wiebe2005annotating}
Janyce Wiebe, Theresa Wilson, and Claire Cardie. 2005.
\newblock Annotating expressions of opinions and emotions in language.
\newblock \emph{Language resources and evaluation}, 39(2):165--210.

\bibitem[{Williams et~al.(2017)Williams, Nangia, and
  Bowman}]{williams2017broad}
Adina Williams, Nikita Nangia, and Samuel~R Bowman. 2017.
\newblock A broad-coverage challenge corpus for sentence understanding through
  inference.
\newblock \emph{arXiv preprint arXiv:1704.05426}.

\bibitem[{Wu et~al.(2017)Wu, Manmatha, Smola, and Krahenbuhl}]{wu2017sampling}
Chao-Yuan Wu, R~Manmatha, Alexander~J Smola, and Philipp Krahenbuhl. 2017.
\newblock Sampling matters in deep embedding learning.
\newblock In \emph{Proceedings of the IEEE International Conference on Computer
  Vision}, pages 2840--2848.

\bibitem[{Wu et~al.(2018)Wu, Xiong, Yu, and Lin}]{wu2018unsupervised}
Zhirong Wu, Yuanjun Xiong, Stella~X Yu, and Dahua Lin. 2018.
\newblock Unsupervised feature learning via non-parametric instance
  discrimination.
\newblock In \emph{Proceedings of the IEEE conference on computer vision and
  pattern recognition}, pages 3733--3742.

\bibitem[{Wu et~al.(2020)Wu, Wang, Gu, Khabsa, Sun, and Ma}]{wu2020clear}
Zhuofeng Wu, Sinong Wang, Jiatao Gu, Madian Khabsa, Fei Sun, and Hao Ma. 2020.
\newblock Clear: Contrastive learning for sentence representation.
\newblock \emph{arXiv preprint arXiv:2012.15466}.

\bibitem[{Xu et~al.(2017)Xu, Xu, Wang, Zheng, Tian, and Zhao}]{xu2017self}
Jiaming Xu, Bo~Xu, Peng Wang, Suncong Zheng, Guanhua Tian, and Jun Zhao. 2017.
\newblock Self-taught convolutional neural networks for short text clustering.
\newblock \emph{Neural Networks}, 88:22--31.

\bibitem[{Yin and Wang(2016)}]{yin2016model}
Jianhua Yin and Jianyong Wang. 2016.
\newblock A model-based approach for text clustering with outlier detection.
\newblock In \emph{2016 IEEE 32nd International Conference on Data Engineering
  (ICDE)}, pages 625--636. IEEE.

\bibitem[{Zhang et~al.(2021)Zhang, Nan, Wei, Li, Zhu, McKeown, Nallapati,
  Arnold, and Xiang}]{zhang-etal-2021-supporting}
Dejiao Zhang, Feng Nan, Xiaokai Wei, Shang-Wen Li, Henghui Zhu, Kathleen
  McKeown, Ramesh Nallapati, Andrew~O. Arnold, and Bing Xiang. 2021.
\newblock \href {https://doi.org/10.18653/v1/2021.naacl-main.427} {Supporting
  clustering with contrastive learning}.
\newblock In \emph{Proceedings of the 2021 Conference of the North American
  Chapter of the Association for Computational Linguistics: Human Language
  Technologies}, pages 5419--5430, Online. Association for Computational
  Linguistics.

\bibitem[{Zhang and LeCun(2015)}]{zhang2015text}
Xiang Zhang and Yann LeCun. 2015.
\newblock Text understanding from scratch.
\newblock \emph{arXiv preprint arXiv:1502.01710}.

\end{thebibliography}
\bibliographystyle{acl_natbib}

\appendix
\begin{table*}[htbp]
  \begin{center}
    \begin{tabular}{lcccccccc}
      & \textbf{MR}& \textbf{CR}&\textbf{SUBJ}& \textbf{MPQA}&\textbf{SST}& \textbf{TREC}&\textbf{MRPC}& \textbf{Avg.} \\
      \cline{2-9}
      BERT$_{\text{distil} \cdot \text{base}}$ &80.0&	85.7&	95.0&	88.9&	85.4&	\textbf{90.6}&	74.1&	85.7 \\
      SBERT$^{\diamondsuit}_{\text{distil} \cdot \text{base}}$&\textbf{80.9}&	\textbf{88.2}&	92.7&	89.5&	\textbf{87.4}&	83.6&	\textbf{75.2}&	85.4\\
      SimCSE$^{\spadesuit}$&-&-&-&-&-&-&-&- \\ 
      \hline 
      \textbf{PairSupCon}$_{\text{distil} \cdot \text{base}}$ &80.6&	87.8&	\textbf{94.7}&	\textbf{89.9}&	86.5&	89.0&	74.4&	\textbf{86.1}
      \\ \\
      BERT$_{\text{base}}$ &81.5&	86.7&	95.2&	88.0&	85.9&	\textbf{90.6}&	73.7&	86.0\\
      SBERT$^{\diamondsuit}_{\text{base}}$ &\textbf{83.6}&	\textbf{89.4}&	94.4&	89.9&	\textbf{89.0}&	89.6&	\textbf{76.0}&	\textbf{87.4} \\
      SimCSE$^{\spadesuit}$&82.9 &89.2 &94.8 &89.7 & 87.3 &88.4 &73.5&86.5  \\
      \hline 
      \textbf{PairSupCon}$_{\text{base}}$ &82.7&	88.8&	\textbf{95.2}&	\textbf{90.3}&	87.6&	88.8&	74.3&	86.8
      \\ \\
     BERT$_{\text{large}}$ &84.3&	89.2&	95.6&	86.9&	89.3&	91.4&	71.7&	86.9 \\
     SBERT$^{\diamondsuit}_{\text{large}}$ &84.6&	\textbf{90.9}&	94.5&	90.2&	\textbf{90.7}&	86.8&	\textbf{76.6}&	87.7 \\
      SimCSE$^{\spadesuit}$&\textbf{85.4} &90.8& 95.3& 90.3& 90.3& 90.6& 76.3& 88.4 \\
      \hline 
      \textbf{PairSupCon}$_{\text{large}}$ &84.5 &	90.1&	\textbf{95.7}&	\textbf{90.6}&	\textbf{90.7}&	\textbf{92.4}&	75.3&	\textbf{88.5} \\
    \end{tabular}
    \caption{Transfer learning results evaluated on SentEval. $\diamondsuit$ and $\spadesuit$: results evaluated on the checkpoints provided by \citet{reimers-2019-sentence-bert} and \citet{gao2021simcse}, respectively. 
    }
    \label{tab:senteval_compare_full_supervision}
  \end{center}
\end{table*}

\section{Implementation}
\label{appendix:implementation}
We use the pre-trained BERT models as the backbone. The pairwise entailment and contradiction classification loss is optimized with a linear layer of size $3d\times 2$, where $d$ denotes the dimension of the sentence representations. We choose a two-layer MLP with size ($d\times d$, $d\times 128$) to optimize the instance discrimination loss. We use Adam \citep{KingmaB14} as our optimizer, with a batch size of 1024 and a constant learning rate of $5e$-04. We use a smaller learning rate $5e$-06 for updating the transformers. We train our models for three epochs and evaluate all downstream tasks at the end of training. Same as \citet{reimers-2019-sentence-bert}, we use mean pooling to obtain sentence representations for both training and evaluation. For fair comparison, we use the default setting, \ie [CLS] representation, to evaluate SimCSE \citep{gao2021simcse}.

\section{Short Text Clustering Datasets}
\label{appendix:cluster_data_stats}
    \paragraph{SearchSnippets} is extracted from web search snippets, which contains 12340 snippets associated with 8 groups \citet{phan2008learning}.
     \paragraph{StackOverflow} is a subset of the challenge data published by Kaggle\footnote{https://www.kaggle.com/c/predict-closed-questions-on-stackoverflow/download/train.zip}, where 20000 question titles associated with 20 different categories are selected by \citet{xu2017self}.

     \paragraph{Biomedical} is a subset of PubMed data distributed by BioASQ\footnote{http://participants-area.bioasq.org}, where 20000 paper titles from 20 groups are randomly selected by \citet{xu2017self}.

     \paragraph{AgNews} is a subset of news
titles \citep{zhang2015text}, which contains 4 topics selected by \citet{rakib2020enhancement}. 
     \paragraph{Tweet} consists of 89 categories with 2472 tweets in total \citep{yin2016model}.

     \paragraph{GoogleNews} contains titles and snippets of 11109 news articles related to 152 events \citep{yin2016model}. We name the full dataset as \textit{GoogleNews-TS} while,  \textit{GoogleNews-T} and \textit{GoogleNews-S} are obtained by extracting the titles and the snippets,  respectively. 

\section{Leveraging STS-Benchmark as the Development Set}
\label{appendix:sts_dev} 
To better understand the underlying causes of the performance gap between PairSupCon and SimCSE on STS, we also train PairSupCon by using STS-Benchmark as the development set. We summarize the corresponding evaluation results in Table \ref{tab:STS_dev}, which indicates that PairSupCon does not benefit from leveraging the STS-Benchmark. We thereby hypothesize the performance gain of SimCSE is mainly attributed by merging the entailment and contradiction discrimination into the instance-wise contrastive learning objective. 

On the other hand, as discussed in Section \ref{subsec:clustering}, SimCSE can be roughly interpreted as an uni-directional instance-wise contrastive learning. In contrast, PairSupCon utilizes the training data in a more efficient way through a bidirectional instance discrimination loss, and hence achieves more obvious performance gain on the clustering tasks. Therefore, developing a better instance discrimination based sentence representation learning objective by incorporating the strengths of both SimCSE and PairSupCon could be a promising direction. 
 
\begin{table*}[htbp]
  \begin{center}
    \begin{tabular}{lcccccccc}
      \textbf{PairSupCon}& \text{STS12}& \text{STS13}&\text{STS14}& \text{STS15}&\text{STS16}& \text{SICK-R}&\text{STS-B}& \text{Avg.} \\
      \cline{1-9}
      \text{DistilBase}$_{\text{w/ STS-B}}$ &73.4 &	83.8	&78.2	&84.2	&81.9	&77.7	&82.2	&80.2 \\ 
      \text{DistilBase}$_{\text{w/o STS-B}}$ &\text{73.6}&	\text{84.1}&	\text{78.3}&	\text{84.4}&	\text{81.8}&	\text{77.8}&	\text{82.3}&	\text{80.3}
      \\ \\
    \text{BertBase}$_{\text{w/ STS-B}}$ & 74.2&	84.5&	79.0&	84.7&	81.4&	78.9&	82.6&	80.8 \\
      \text{BertBase}$_{\text{w/o STS-B}}$ &74.3&	84.4&	79.0&	84.8&	\text{81.4}&	79.0&	82.6&	80.8
      \\ \\
     \text{BertLarge}$_{\text{w/ STS-B}}$& 74.3&	85.7&	79.1&	84.8&	81.7&	80.2&	83.3&	81.3 \\
      \text{BertLarge}$_{\text{w/o STS-B}}$& 74.2&	85.8&	79.5&	85.5&	\text{81.7}&	80.4&	83.7&	81.5 \\
    \end{tabular}
    \caption{Investigation on the effectiveness of leveraging STS-Benchmark as the development set for training PairSupCon. The training of our proposed PairSupCon model is fully blind to the downstream tasks.  This table indicates that, different from SimCSE \citep{gao2021simcse}, no obvious performance gain is attained by PairSupCon when using STS-B as the development set. 
    }
    \label{tab:STS_dev}
  \end{center}
\end{table*}

\section{Transfer Learning}
\label{appendix:transfer_learning}
To provide a fair and more comprehensive comparison with the existing work, we also evaluate PairSupCon on the seven transferring tasks using the SentEval toolkit \citep{conneau2018senteval}. We follow the widely used evaluation protocol, where a logistic regression classifier is trained on top of the frozen representations, and the testing accuracy is used as a measure of the representation quality. We report the evaluation results in Table \ref{tab:senteval_compare_full_supervision}. As we can see, the performance gap between different models are small, yet still not consistent across different tasks. As discussed in Section \ref{subsec:transfer_learning}, one possible explanation is that the transfer learning tasks do not present enough complexities to discriminate the performance gap between different approaches, since most tasks are binary classification with a large amount of labeled training examples. 

Although we observe more obvious performance gap by extending the evaluation to the setting of few-shot learning (Table \ref{tab:senteval_fsl_compare}), there is no consistent performance gain across different tasks attained by any specific model investigated in this paper. Moving forward, having more complex and diverse datasets to evaluate the transfer learning performance could better direct the development of universal sentence representation learning.

\end{document}